\documentclass[10pt,twocolumn,letterpaper]{article}

\usepackage[pagenumbers]{cvpr}      

\usepackage{graphicx}
\usepackage{amsmath}
\usepackage{amssymb}
\usepackage{booktabs}
\usepackage{dsfont}
\usepackage{tabu}
\usepackage[table]{xcolor}
\usepackage{multirow}
\usepackage{soul}
\usepackage{subcaption}

\usepackage{array}
\newcolumntype{K}[1]{>{\centering\arraybackslash}p{#1}}

\usepackage[subtle]{savetrees}

\usepackage{overpic}
\newcommand{\sgmcolorval}{0.95}
\definecolor{sgmgray}{rgb}{\sgmcolorval, \sgmcolorval, \sgmcolorval}

\usepackage[pagebackref,breaklinks,colorlinks]{hyperref}

\usepackage[capitalize]{cleveref}
\crefname{section}{Sec.}{Secs.}
\Crefname{section}{Section}{Sections}
\Crefname{table}{Table}{Tables}
\crefname{table}{Tab.}{Tabs.}

\begin{document}

\title{Learning to Name Classes for Vision and Language Models}

\author{Sarah Parisot \qquad Yongxin Yang \qquad Steven McDonagh \\
Huawei Noah's Ark Lab \\
{\tt\footnotesize \{sarah.parisot, yongxin.yang, steven.mcdonagh\}@huawei.com}}

\maketitle

\begin{abstract}
Large scale vision and language models can achieve impressive zero-shot recognition performance by mapping class specific text queries to image content. Two distinct challenges that remain however, are high sensitivity to the choice of handcrafted class names that define queries, and the difficulty of adaptation to new, smaller datasets. Towards addressing these problems, we propose to leverage available data to learn, for each class, an optimal word embedding as a function of the visual content. By learning new word embeddings on an otherwise frozen model, we are able to retain zero-shot capabilities for new classes, easily adapt models to new datasets, and adjust potentially erroneous, non-descriptive or ambiguous class names. We show that our solution can easily be integrated in image classification and object detection pipelines,  yields significant performance gains in multiple scenarios and provides insights into model biases and labelling errors.  
\end{abstract}

\section{Introduction}
\label{sec:intro}

The introduction of large-scale vision and language pre-training techniques has led to significant breakthroughs in object recognition, notably for tasks where data efficiency forms a crucial component (\eg zero-shot scenarios). The key idea is to learn a mapping between image and text spaces such that a latent image representation will be close to sentences that describe the image content, in representation space~\cite{radford2021learning,minderer2022simple}. Learning from image-specific descriptions, rather than a fixed and common class label, with limited semantic meaning, yields models with very strong representational power~\cite{radford2021learning}. 
Such multi-modal approaches address one key shortcoming of visual-only recognition methods; they provide an ability to seamlessly expand the model's output space without the requirement of additional training (\ie open-set recognition). In contrast, visual-only classification and object detection methods have historically relied on training a linear classification layer in order to identify objects of interest, resulting in the undesirable consequence that any introduction of new classes (with or without available data) is very challenging. A plethora of complex mechanisms have been proposed, towards addressing this problem~\cite{joseph2021towards,geng2020recent}. Vision and language multi-modal approaches alternatively perform classification by computing a similarity between image features and a set of textual features, corresponding to text descriptions 
from the classes of interest. As the text component constitutes an input, such models naturally enable a route to identify new categories, simply by modifying the queried text, yielding very strong zero-shot and open-set performance~\cite{radford2021learning, kamath2021mdetr}.

\begin{figure}[t]
\centering 
\hfill
\subfloat[][\centering Ambiguous class names]{
\begin{overpic}[width=0.45\textwidth]{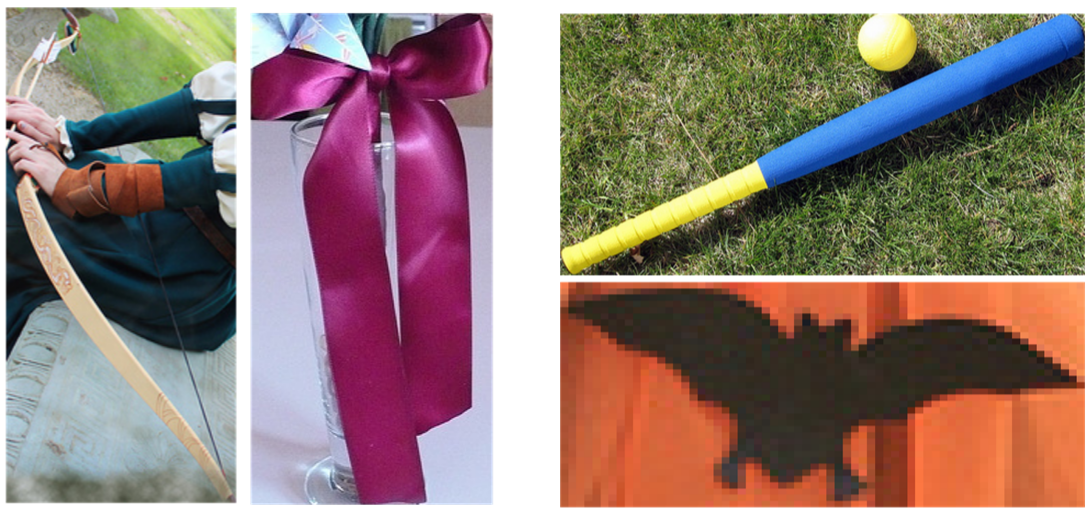}
\put(12,48){\footnotesize {Both named ``\emph{bow}''}}
\put(63,48){\footnotesize {Both named ``\emph{bat}''}}
\end{overpic}
\label{fig:teaser:a}
}
\hfill
\vspace{6.5mm}
\subfloat[][\centering Technical class names]{
\begin{overpic}[width=0.475\textwidth]{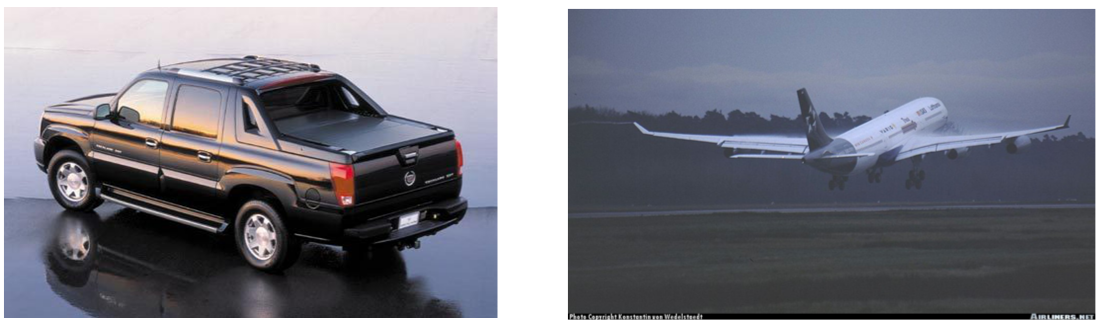}
\put(-2,30){\scriptsize {\shortstack{Class name: \\ ``\emph{2007 Cadillac Escalade EXT Crew Cab}''}}}
\put(68,30){\scriptsize {\shortstack{Class name: \\ ``\emph{A3400-200}''}}}
\end{overpic}
\label{fig:teaser:b}
}
\hfill\null
\caption{\small Vision and language recognition models are highly sensitive to class prompts. Ambiguous or highly technical class names can lead to poor classification performance. Rather than adjusting class names directly, current solutions rely on prompt context to reduce ambiguity, \eg:~`\emph{a photo of a bow, a type of weapon}'.}
\label{fig:teaser}
\end{figure}

We draw attention to the fact that such performant multi-modal models suffer from two remaining key limitations. Firstly, performance is highly sensitive to class-prompt textual content; both in terms of ($i$) the selected class name and ($ii$) the surrounding sentence context. Class name sensitivity can be illustrated by considering class homonyms; words that have the same spelling but different meanings and origins (Figure \ref{fig:teaser:a}). These lead to ambiguous queries that negatively effect performance and mitigation necessitates time consuming prompt engineering. A further example of class name sensitivity can be demonstrated by classes exhibiting highly technical characterisations (Figure \ref{fig:teaser:b}). Poor performance is typically observed for datasets exhibiting such complex naming strategies~\cite{radford2021learning}. We hypothesise that lack of clear semantic meaning leave it difficult for multi-modal models to obtain valuable embeddings. 

Secondly; model adaptation to new data, that becomes available, is not straightforward -- especially when datasets are small. A natural solution is to use the existing representation layers and learn a new classification layer (\ie linear probing). However this strategy must sacrifice the flexibility of text input, for potential subsequent adaptations~\cite{radford2021learning}. Alternatives consider replacing text queries with image queries~\cite{minderer2022simple}, or learning optimal prompt contexts from the new data~\cite{zhou2022learning,zhou2022conditional,du2022learning}. 
The former can struggle to leverage larger amounts of available data successfully as image queries are averaged over all available class specific data. The latter alternatively updates a fixed set of parameters, which can lead to forgetting issues under sequential adaptation. Furthermore, the approach fails to learn a new class representation or mapping from the new data, as the focus lies solely with the prompt context. Finally; while context learning is highly successful in simple classification tasks, it becomes more challenging for object detection purposes and requires more complex training mechanisms~\cite{du2022learning}. We conjecture that this may be attributed to the fact that context then depends on the spatial {region} of the image, in addition to object type (\ie foreground or background). 

These considerations lead us to propose a complementary solution to prompt learning, which can be successfully leveraged for both classification and detection problems, without any task-specific tailoring or requirements. We build upon ideas from the field of text to image generation, and propose to leverage the concept of textual inversion~\cite{gal2022image} to adapt models to new data and thus address suboptimal, and potentially wrong, labels (\eg incorrect class names, mislabelled data). The crux of our idea is to supersede manual word engineering, describing categories of interest, by leveraging available data to learn optimal word embeddings, as a function of the visual content. This simple solution affords several advantages: 1) in contrast to linear probing, open-set capabilities are maintained; learned class representations remain in text space, 2) our approach learns class specific parameters, therefore preventing forgetting under sequential training, 3) direct applicability to any classification or object detection technique that utilises textual input, 4) interpretability properties, with regards to learned class names.  

Experimentally, we realise our ideas in conjunction with two contemporary models; for image classification~\cite{radford2021learning} and object detection~\cite{minderer2022simple}. Given a pre-trained model and (potentially previously unseen, additional) training dataset, we introduce a new set of word embeddings in the model text branch, each corresponding to a class of interest. Following context tuning methods \cite{zhou2022learning, zhou2022conditional}, embeddings are learned using standard losses on an otherwise frozen model. Comprehensive experiments across thirteen 
classification and detection datasets help to evidence that learning optimal class names can significantly enhance model performance, maintain open-vocabulary properties, successfully enable continual model adaptation, and improve performance for rare and longtail classes. Last but not least, we demonstrate the method's potential for interpretability, notably with regards to mislabelling and inadequately chosen original class names. 

Our main contributions can be summarised as:
\begin{itemize}
\item A simple proposition for data-efficient adaptation of large vision-language models: optimisation of class names from visual content that retains attractive open-vocabulary properties.
\item Performance improvements \emph{complementary} to those of prompt tuning; consistent experimental gains across 13 classification and detection datasets. Strong sequential adaptation, open-vocabulary and long tail performance. 
\item Model interpretability insights: we visualise class name changes and related semantic meanings. We highlight how model biases and labelling errors can be identified, and how vision-language models suffer from long-tail issues.
\end{itemize}

\section{Related work}

\paragraph{Vision and language models.}

Early vision and language multi-modal alignment work tended to enable learning by encoding distinct modalities individually and independently. Model branch outputs would then be connected by losses~\cite{simonyan2014very}.
Cross-modality alignment has been performed using techniques including metric learning~\cite{frome2013devise}, multi-label classification~\cite{joulin2016learning} and transformers~\cite{zheng2020cross}. A contemporary alternative approach pertains to so called ``vision-language'' models~\cite{furst2021cloob,jia2021scaling,li2021supervision,radford2021learning} that recently gain popularity and instead are designed to jointly learn modality encoders; sharing information between modalities to the benefit of the global model. A representative approach is CLIP~\cite{radford2021learning}, which trains two neural network-based encoders on a very large image-caption dataset using a contrastive cross-entropy loss, to match pairs of images and texts, and shows impressive zero-shot image recognition performance. This seminal work has been extensively investigated in recent literature, and several improvements have been proposed ~\cite{furst2021cloob,jia2021scaling,LIT}.

Open vocabulary object detection was originally explored in a two-step manner~\cite{zareian2021open}: firstly learning a mapping between the two modalities using a large-scale dataset, then leveraging the obtained encoders and mapping layers to learn an object detector that can perform classification by computing a distance to class names' text embeddings. More recent methods fully integrate vision-language mappings into the object detection training process, providing more flexible and adaptable multi-modal detection solutions~\cite{kamath2021mdetr,cai2022x,minderer2022simple,vild,li2021grounded, maaz2021multi}. Alternative methods generate a multi-modal joint representation via a fusion block~\cite{kamath2021mdetr, maaz2021multi}, while contemporary solutions adopt a CLIP-like parallel training strategy~\cite{cai2022x,minderer2022simple,vild}, facilitating querying of large numbers of classes. Similarly to the classification setting, multi-modal detection models achieve impressive zero-shot performance.  

\noindent\textbf{Prompting methods.}
Prompt learning aims to exploit the rich knowledge contained in large pre-trained language models (\eg GPT~\cite{radford2019language}), to predict new downstream tasks. 
The idea prescribes that we can reformulate these new tasks to ``look more like'' those solved during the original pre-training stage, through the help of an appropriate textual prompt, such that the model can be used to predict the desired output without any additional training
~\cite{radford2019language,schick2020s,gao2020making,liu2021pre}. 
The disadvantage with such methods, is the necessity for \emph{prompt engineering}; searching for the most suitable prompt to allow the language model to solve the desired task.
Recent work has looked to learn optimal prompts \cf manually designing  them. Work on continuous prompt learning methods~\cite{lester2021power,li2021prefix,zhong2021factual} can be considered closely related to our work, where the main idea is to convert the prompt into a set of continuous vectors that can be optimised, with respect to an objective function, in end-to-end manner. 

The benefits of the prompting strategy have recently started to be explored in computer vision~\cite{ju2022prompting,rao2022denseclip,yao2021cpt,zhang2022pointclip,zhou2022learning}. CoOp~\cite{zhou2022learning} can be considered the first work to bring continuous prompt learning to the vision domain, towards offering efficient solutions for adaptation of pre-trained vision-language models to downstream vision tasks. CoOp was recently extended by~\cite{zhou2022conditional} where they propose to learn an additional light-weight neural network to generate, for each image, an input-conditional token that yields an image-specific dynamic prompt, outperforming CoOp's static prompts. This comes at the price of efficiency however, as the proposed extension is substantially both more memory and computationally expensive. We note that~\cite{zhou2022learning,zhou2022conditional} facilitate model adaptation to new domains in a highly parameter efficient manner, yet may still struggle in settings where new classes (and their names) widely differ from that which the model has initially learned. A more complex context prompting strategy has been recently proposed, specifically for detection tasks~\cite{du2022learning}; integrating a background specific loss function, as well as a context dependent scheme in the prompt learning process.  

Crucially, for all of these methods, classification is still performed with respect to a handcrafted class name, making accuracy highly dependent on the quality of chosen names. In certain settings, finding an accurate semantic description of a class can be challenging, and a poor choice can impair the model's performance. To the best of our knowledge, this limitation has not been addressed yet in classification and object detection settings. 

\begin{figure}[t]
\centering
\includegraphics[width=\linewidth]{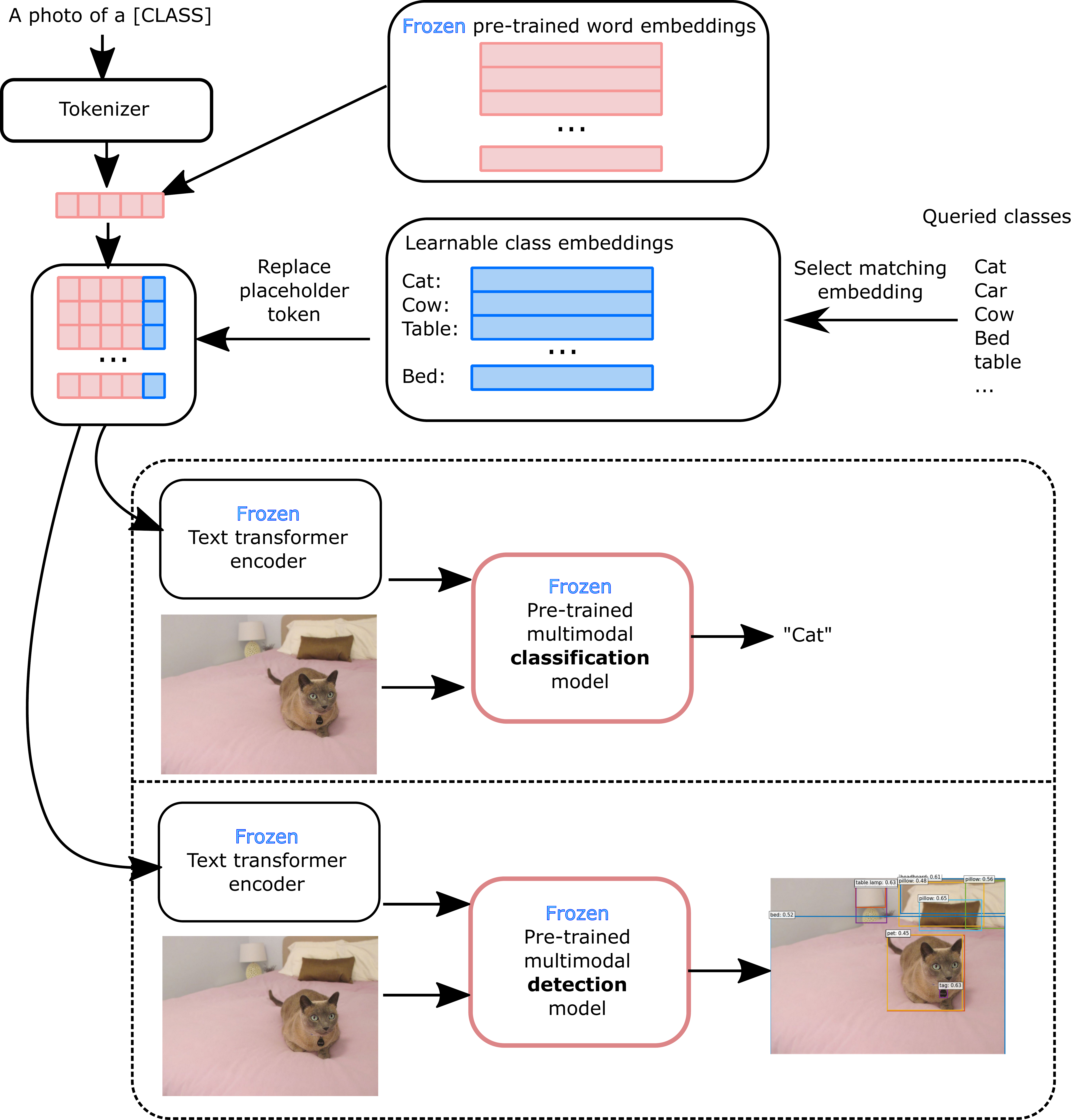}
\caption{\small Overview of our method. Classes of interest are mapped to a set of learnable word embeddings, and integrated within the input sentence representation. These embeddings can be learned for classification and object detection tasks using standard losses, and used in conjunction with pre-trained word embeddings in open-vocabulary settings.}
\label{fig:overview}
\end{figure}

\noindent\textbf{Textual Inversion.}
Textual inversion was recently introduced in~\cite{gal2022image} for text to image generative diffusion models, as a means to condition image generation towards incorporating specific objects or visual components. Given a set of images, all displaying the same object in diverse settings, a specific word embedding is learned so as to ensure the model is able to reconstruct images, sampled from this small set. Simple prompt contexts are used (\eg `a photo of') to ensure a simple image of the object is generated. After training, the model displays impressive ability to integrate specific objects in generated images, via inputting learned words at specific locations in the input sentence. In the solution proposed by~\cite{gal2022image}, each word is learned individually.

\section{Methodology}
\label{sec:method}

We consider a vision and language object recognition model pre-trained on a large-scale dataset. Following the multi-modal architecture, the model takes as input an image-text pair $\mathbf{x}=\{I,T\}$, where the text provides a list of $N$ candidate classes $t_i$ to recognise; \mbox{$T=[t_1,t_2,\ldots,t_{N}]$}. We assume the $i$-th class specific text $t_i$ to be in the format: \mbox{$t_i= [\text{prompt prefix}] + [\text{CLASS}] + [\text{prompt suffix}]$}, where prompt prefixes and suffixes are shared across all classes. To obtain text feature representations, a tokenizer converts each sentence into a series of tokens such that each token maps to a pre-trained word embedding $\mathbf{e_k}$. Embeddings are then concatenated and used as input to a text transformer. Our final text representation is the transformer output for the End Of Sequence (EOS) token $f(T) = f_i(t_i)$, where $i \in \{1,\ldots,N\}$.

\subsection{Baseline multi-modal models}
\label{sec:method:baselines}

\paragraph{Multi-modal Image classification.}
In the context of image classification~\cite{radford2021learning}, our objective is to assign a specific category to the image as a whole. Image inputs are processed in parallel to the text input through an image encoder $g(\cdot)$, yielding representation $g(I)$. The classification task is then simply performed by evaluating the cosine similarity between image and text embeddings, and selecting the class with the highest similarity.

\noindent\textbf{Multi-modal Object detection.}
There are two main contemporary strategies for vision and language based object detection. The first (simpler) strategy is similar to multi-modal classification, wherein image and text information are only combined latterly at the bounding box classification stage~\cite{cai2022x, minderer2022simple, vild}. In this setting, the image input is processed through an object detector entirely disjointly from the text modality. The detector outputs a set of coordinates and a class embedding for each bounding box candidate. We then compute class embedding similarity with text representations $f(T)$ following the multi-modal classification approach to assign a classification label to each bounding box.

The second class of strategies propose to fuse image and text inputs; either when generating text and image embeddings~\cite{li2022grounded} or via a third encoder $h(g(I),f(T))$ that outputs bounding box coordinates and class embeddings~\cite{kamath2021mdetr,vild, maaz2021multi}. For this latter fusion strategy, classification is performed over sentence tokens rather than across sentences, with an aim of associating bounding boxes with a keyword. This technique is typically more 
computationally expensive than the first method class as it requires computation of new image features, for each queried class. 

\subsection{Learning optimal class names}
\label{sec:method:optclas}
We conjecture that the performance of pre-trained recognition models is intrinsically linked to the suitability of the class names, chosen to represent certain visual categories. Selecting class names that do not reflect an object's visual appearance can substantially alter recognition performance. We thus propose to remove the model's sensitivity to hand crafted class names, by \emph{learning class specific word embeddings from image content}. An overview of the method is provided in Figure~\ref{fig:overview}.

We consider the matrix of pre-trained word embeddings $E \in \mathds{R}^{V{\times}F}$, where $V$ is the vocabulary size, and $F$ the feature dimension. We propose to extend this matrix with a set of new, learnable word embeddings $E^l \in \mathds{R}^{N{\times}F}$, where each row of $E^l$ corresponds to a class specific learnable embedding and $N$ pertains to the total number of dataset classes.  In this setting, we replace the class name in each text query with a class specific placeholder \mbox{$t_i= [\text{prompt prefix}] + [\text{pl}^i] + [\text{prompt suffix}]$}. While the tokenizer maps context prompts to matrix $E$, our placeholder maps to its corresponding class embedding $\mathbf{e}^l_i$. This new class query is used as input to the text transformer encoder, and the remainder of the recognition model is unchanged. We note that it is possible to map one category to a series of $m$ word embeddings, using queries of the type $t_i= [\text{prompt prefix}] + [\text{pl}^i_1] + \cdots 
+ [\text{pl}^i_m] + [\text{prompt suffix}]$. 

\noindent\textbf{Training.} During training, we freeze the entire pre-trained model (including pre-trained word embeddings) and learn our new set of word embeddings $E^l$ with the standard losses used to train the base model (\eg cross entropy loss for classification, and bounding box regression losses for detection). We identify three key scenarios where our approach can be beneficial: 1) adapting the pre-trained model on a new dataset with a new set of classes; 2) adjusting suboptimal class names on the same dataset used to train the model; 3) continual adaptation of a pre-trained model, where new classes are available after a first training stage.  

\noindent\textbf{Inference.} At inference time, we follow the process described in Section~\ref{sec:method:baselines}. Categories whose names were learned are mapped to their new embeddings $e^l \in E^l$. For unseen object categories (\ie not in our new dataset) we use their original embeddings, $e \in E$, from the pre-trained model. As a result, our solution maintains the ability to perform open-vocabulary recognition.

\section{Experiments}
\label{sec:exp}

We evaluate the impact of learning class names for the tasks of image classification and object detection. We consider four settings where our strategy can be beneficial: 

\noindent\textbf{Task 1: Model adaptation to new datasets}: considering a new dataset of interest, we adapt our pre-trained model to this new set of potentially unseen classes by learning optimal class names from the new image content.

\noindent\textbf{Task 2: Post-training class name adjustment}: the model is first fine-tuned on our target dataset, then class names are optimised in a subsequent learning stage. This setting is more valuable for object detection where fine-tuning can have important impact on bounding box localisation.

\noindent\textbf{Task 3: Open-vocabulary recognition:} open-vocabulary is one of the key advantages of vision-language models, and maintaining strong performance is crucial. Evaluation of the compatibility between our learned word embeddings, and standard pre-trained word embeddings can provide a measure of strength on this axis. We adapt the model to only a subset of classes from our new dataset and measure overall performance on all classes, such that inference combines learned and text based word embeddings. 

\noindent\textbf{Task 4: Continual model adaptation:} to measure the model ability for sequential learning of new word embeddings, we first adapt the model to a subset of classes, then learn the remaining class names in a second, separate, training stage.

We highlight that we do not seek state of the art performance, as this is intrinsically linked to the pre-trained model's base performance, pre-training data, and architecture. Rather, our objective is to demonstrate the potential of learning class names to improve model performance, while maintaining zero-shot properties. We therefore perform experiments on models with simple multi-modal mechanisms, use a single word embedding per class, and a standard prompt: \mbox{($ \text{prompt prefix}=\text{`{a photo of a}'} $; $\text{prompt suffix}=\text{`.'}$)}, in all experimental settings unless specified otherwise.

\subsection{Image classification: CLIP model}
\label{sec:exp:class}

To evaluate our method on classification tasks, we employ 
CLIP~\cite{radford2021learning} as our base pre-trained model with a vision transformer ViT-B/16~\cite{dosovitskiy2020image} fulfilling the image encoder $g(\cdot)$ role. Here, we seek to evaluate method performance in previously described \textbf{Tasks} \textbf{1}, \textbf{3} and \textbf{4}; \ie model adaptation, open-vocabulary performance, and continual adaptation. 

\noindent\textbf{Datasets.} We follow the setting proposed in~\cite{zhou2022learning,zhou2022conditional}. We adapt the CLIP model to a selection of eleven diverse datasets, focusing on different types of recognition tasks: standard image classification (ImageNet~\cite{deng2009imagenet}, Caltech 101~\cite{fei2004learning}), fine-grained classification (Stanford cars~\cite{krause20133d}, Oxford Pets~\cite{parkhi2012cats}, Flowers 102~\cite{nilsback2008automated}, Food 101~\cite{bossard2014food} and FGVC Aircraft~\cite{aircraft}), scene recognition (SUN 397~\cite{sun397}), action recognition (UCF 101~\cite{ucf101}), texture classification (DTD \cite{dtd}) and satellite image recognition (Eurosat~\cite{eurosat}).

\noindent\textbf{Training details.} We adopt a set of parameters similar to~\cite{zhou2022learning}. For each dataset, we learn class names with a batch size of $32$, over $200$ epochs ($50$ for ImageNet), SGD optimiser with a learning rate of $2e^{-4}$ and cosine scheduling, with one warm up epoch. We initialise our learnable word embeddings using the average embedding of all word tokens associated with hand-crafted class names.  For each dataset, 16 training images per category are sampled, and we report mean statistics over three random seeds. 

\noindent\textbf{Baselines.} The method we directly compare to is the zero-shot CLIP model, as we aim to evaluate the benefits of learning class names. We additionally report classification accuracy for the CoOp model~\cite{zhou2022learning}, which learns optimal prompt context with fixed handcrafted class names. We retrain CoOp in our settings of interest and use four context words, initialised with `{a photo of a}', as in \cite{zhou2022conditional} (this differs from the 16 words used in ~\cite{zhou2022learning}). This allows to compare advantages and disadvantages of learning class names over prompt context, and evaluate any complementary nature between the strategies. We report the best possible CLIP performance with engineered prompts, and report our results with a standard prompt, as well as using the same engineered prompts as CLIP. We note that four of the datasets (Stanford cars, ImageNet, Caltech $101$ and Sun $397$) do not have engineered prompt, and we provide all dataset-specific prompts in the supplementary materials.  

\begin{table}[t]
\centering
\resizebox{0.75\linewidth}{!}{\begin{tabular}{l||l|l|l|l}
\toprule
Class group & CLIP* & CoOp & Ours & Ours* \\
\midrule
Base classes & 69.4 & 82.6 & 82.7 & 82.8  \\
New classes & \bf74.2 & 64.1 & 73.1 & \bf74.2  \\
All classes & 65.2 & 65.7 & 66.5 & \bf68.7 \\
\bottomrule
\end{tabular}}
\caption{\small CLIP model: Base to new classification accuracy. Results on Base classes show method ability to adapt to new data (Task \textbf{1}). Results on New and All classes show open vocabulary ability (Task 3). * Manually engineered prompt templates.}
\label{tab:classif:base2new}
\end{table}

\noindent\textbf{Model adaptation (Task \textbf{1}) and Open vocabulary (Task \textbf{3}).} We adopt the base-to-new generalisation setting proposed in~\cite{zhou2022conditional}. For each dataset, we equally split classes into a set of `base' and `new' classes. We use base classes for training and evaluate performance on new classes in a zero-shot manner. Additionally, to measure open-vocabulary performance, we report performance on `all' classes, \ie a mixture of learned and handcrafted class names. Results are reported in Table~\ref{tab:classif:base2new}. 
Results on base classes highlight our ability to perform model adatpation (Task \textbf{1}) by learning class names only. For this task, we observe that our strategy can achieve equivalent gains as context tuning on base classes (82.6\% (CoOP) \vs 82.8\% (Ours) average accuracy).  However, our approach guarantees that zero-shot CLIP performance is maintained on new classes, as our learned parameters are class specific. Our open-vocabulary experiments (Task \textbf{3}) show that we are able to maintain the CLIP model's ability to expand the output set, and to introduce new classes from text alone (CLIP and Ours 74.2\%, new classes). Experiments on all classes confirm that we are able to use a mixture of learned and new classes successfully, outperforming CoOp (65.7\% CoOp, 68.7\% Ours, all classes). We further note sensitivity to context prompts, as substantial gains are obtained using engineered prompts (+2.2\% Ours, all classes). This is to be expected, as new class performance can be strongly impacted by prompts, and highlights the importance of prompt context tuning in this setting. 

\begin{table}[t]
\centering
\resizebox{0.75\linewidth}{!}{\begin{tabular}{l||l|l|l|l}
\toprule
Class group & CLIP* & CoOp & Ours & Ours*  \\
\midrule
New classes &  74.0 & 84.4 & \bf85.9 &\bf85.9 \\
All classes &   65.2 & 71.7 & 75.7 & \bf76.3\\
\bottomrule
\end{tabular}}
\caption{\small CLIP model: Classification accuracy for continual model adaptation (Task 4) experiments. * manually engineered prompt templates. }
\label{tab:classif:seq}
\end{table}

\noindent\textbf{Sequential adaptation (Task \textbf{4}).} We extend the generalisation setting to evaluate our ability to perform sequential adaptation; \ie after training on base classes, we subsequently train on new classes with no access to base classes. We measure performance on new classes, as well as all classes after this second training stage. For CoOp \cite{zhou2022learning} experiments, we fine-tune prompt parameters, and adjust learning rate to $2e^{-4}$ which provided best empirical performance in this setting. Results are reported in Table~\ref{tab:classif:seq}. As in the previous setting, performance on new classes is highly similar when performing context or class name learning (84.4\% CoOp, 85.9\% Ours, new classes), achieving strong gains over zero-shot CLIP (10+\%). On all classes, we can see that we achieve significant gains (+4\%, Ours \vs CoOp),
which can be attributed to the fact that class names are (all) learned independently. 
Last but not least, we note a much higher robustness to change in prompt context when all classes are learned; we achieve very similar results when using a standard or engineered template in this setting (+0.6\% \vs +2.2\% in Table \ref{tab:classif:base2new}, all classes).

\begin{table*}[t]
\centering
\resizebox{0.75\linewidth}{!}{\begin{tabular}{l|l|l|l|l|l|l|l}
\toprule
Method &  backbone & pre-training (image) & pre-training (detection) & AP & AP/f & AP/c & AP/r \\
\midrule
\multicolumn{7}{l}{\textbf{Zero-shot models}} &  \\
\midrule
\rowcolor{sgmgray} MDETR \cite{kamath2021mdetr} & DETR& N/A & MDETR data & 6.4 & 9.8  & 3.6 & 1.9 \\
\rowcolor{sgmgray} XDETR \cite{cai2022x} & DETR& N/A & MDETR data, LocNar& 16.4 & 18.8  & 15.2 & 9.6 \\
\rowcolor{sgmgray} GLIP-T \cite{li2021grounded} & SWIN-T & Cap4M & O365,GOLDG & 26.0 & 31.0 & 21.4 & 20.8 \\
\rowcolor{sgmgray}
GLIP-L \cite{li2021grounded} & SWIN-L & Cap24M & O365,FourODs &  \bf37.3 & \bf41.5 & \bf43.3 & \bf28.2 \\
\midrule
\multicolumn{7}{l}{\textbf{Zero-shot OWL-vit base model} }&  \\
\midrule
OWL-vit zero-shot \cite{minderer2022simple} & VIT B/16 & CLIP & O365, VG & 25.3 & 27.0 & 25.0 & 17.6 \\
Prompt learning all \cite{zhou2022learning} & VIT B/16 & CLIP & O365, VG & 24.4 & 25.9 & 24.0 & 17.8 \\
Ours base (Task \textbf{3}) &  VIT B/16 & CLIP & O365, VG & 25.7 & 26.8 & 25.9 & 19.0 \\
Ours sequential (Task \textbf{4}) &  VIT B/16 & CLIP & O365, VG & 26.0 & 26.6 & 25.9 & \bf23.1 \\
Ours all (Task \textbf{1}) & VIT B/16 & CLIP & O365, VG & \bf26.6 & \bf27.0 & \bf27.5 & 19.8 \\
\midrule
\multicolumn{7}{l}{\textbf{OWL-vit base model fine-tuned on LVIS base classes}} \\
\midrule
OWL-vit \cite{minderer2022simple} & VIT B/16 & CLIP & O365, VG &28.8 & 34.4 & 24.7 & 17.7 \\
Prompt learning all \cite{zhou2022learning} & VIT B/16 & CLIP & O365, VG & 31.3 & 34.7 & 29.2 & 22.4 \\
Ours base (Task \textbf{3}) & VIT B/16 & CLIP & O365, VG &33.3 & 35.2 & 33.9 & 20.3 \\
Ours rare (Task \textbf{3}) & VIT B/16 & CLIP & O365, VG &30.4 & 34.5 & 25.4 & \bf 34.5 \\
Ours sequential (Task \textbf{4}) & VIT B/16 & CLIP & O365, VG &34.1 & 35.0 & 33.8 & 30.6 \\
Ours all (Task \textbf{2}) & VIT B/16 & CLIP & O365, VG &\bf 34.5 & \bf 35.6 & \bf 34.4 & 28.5 \\
\midrule
\multicolumn{7}{l}{\textbf{OWL-vit base model fine-tuned on all LVIS classes}} \\
\midrule
OWL-vit \cite{minderer2022simple} & VIT B/16 & CLIP & O365, VG &34.5 & 38.5 & 33.2 & 19.1 \\
\bottomrule
\end{tabular}}
\caption{\small Average precision results on LVIS mini-validation set for all (AP), frequent (AP/f), common (AP/c) and rare (AP/r) classes. Our method uses only 10\% of the LVIS training dataset. State of the art methods (in grey) use different training datasets and architectures of different capacities. They are added for context. \{base, rare, all\} describes which class group is used for training. Sequential means base and rare classes are trained sequentially. Task 1: model adaptation to new data, Task 2: class name learning after fine-tuning, Task 3: Open vocabulary ability (mixture of seen and unseen classes), Task 4: continual adaptation (learning class groups in a seqeuntial manner).}
 \label{tab:det:lvis}
\end{table*}

\subsection{Object detection: OWL-vit model}
\label{sec:exp:det}

For object detection experiments, we evaluate all 4 Tasks. We select OWL-vit~\cite{minderer2022simple} as our base multi-modal detection model due to noted strong zero-shot performance and clear separation between vision and language branches. This allows us to evaluate the impact of class learning more directly, in contrast to methods using complex text and image fusion mechanisms~\cite{li2021grounded}. 
OWL-vit leverages a pre-trained multi-modal classification model (\eg CLIP) with a vision transformer backbone. To adapt this model to the object detection task, the pooling layer that combines vision transformer outputs is removed, and two new modules are learned: a class prediction module and a box prediction module that output class embeddings and bounding-box coordinates, respectively, for all transformer outputs. The OWL-vit model is fine-tuned from the CLIP weights using a mixture of standard object detection datasets. Additionally, a set of engineered CLIP prompt templates are used as class prompt contexts. 

\noindent\textbf{Datasets.} We evaluate the impact of learning class names on two object detection datasets. Firstly, we consider LVIS~\cite{lvis}, a large scale natural image object detection dataset comprising over $100k$ training images and $1203$ classes, characterised by a long tail distribution with `frequent' ($x{>}100$), `common' ($100{\ge}x{>}10$), and `rare' ($10{\ge}x{\ge}1$) class groups, for $x$ images per class. Following prior work~\cite{cai2022x}; we use pre-defined category names, but remove the text in the parentheses, \eg, ``flip-flop (sandal)'' is replaced by ``flip-flop''. We note that this leads to a few instances of classes sharing the same name (\eg ``bow'').

Secondly, we run experiments on CODA 2.0~\cite{li2022coda}, a self-driving dataset focusing on detection of rare, corner case instances. It comprises validation and test sets containing $5k$ images each, and $43$ classes, of which only $29$ appear in the validation set. In addition to the base OWL-vit model, we consider a model fine-tuned on the SODA
dataset~\cite{han2021soda10m} (fully labelled instance), closer to our test domain. Designed for semi-supervised training of self-driving models, SODA comprises six classes that overlap with CODA classes. Following~\cite{li2022coda}, we refer to the \{\emph{pedestrian, cyclist, car, bus, tram, truck, tricycle}\} classes as common classes, and remaining classes as corner-cases.

\noindent\textbf{Training details.} We fine-tune the pre-trained OWL-vit model on LVIS, SODA datasets for 10 epochs, using 7 GPUs (batch size 2/GPU), with learning rates $5e^{-6}$ (main model) and $2e^{-7}$ (text branch) using cosine scheduling and an Adam optimiser. Weight decay is unused and gradient clipping is set to $0.1$. For LVIS fine-tuning, we train two models: one on all classes, and one on base (frequent + common) classes only, thus following the evaluation set-up of~\cite{minderer2022simple}. For CODA 2.0 experiments, models are fine-tuned on the SODA and LVIS datasets, to simulate the setting where corner cases are discovered after training.  \\
For our class learning process, we freeze the entire model aside from new class embeddings. Embeddings are learned for $20$ epochs, using a learning rate of $1e^{-2}$ (cosine schedule), 7 GPUS (batch size 10/GPU). They are initialised with the average token embedding of the corresponding class name. For the LVIS dataset, we learn class names with a balanced subsampling of $10\%$ of the training set.

\begin{table}[t]
\centering
\resizebox{0.9\linewidth}{!}{
\begin{tabular}{l|lllll}
\toprule
Method & AP-Com & AP-A & AR-A & AR-Cor & Sum \\
\midrule
RetinaNet & 0.25 & 0.38 & 0.48 & 0.25 & 1.36 \\
Faster   RCNN & 0.28 & 0.37 & 0.43 & 0.17 & 1.27 \\
Cascade   RCNN & \bf0.30 & \bf0.42 & 0.48 & 0.21 & 1.41 \\
\midrule
\multicolumn{6}{l}{\bf Zero-shot OWL-vit base model} \\
\midrule
OWL-vit zero-shot & 0.10 & 0.20 & 0.44 & 0.45 & 1.19 \\
Ours (Task \textbf{3}) & 0.17 & 0.31 &  0.47 & 0.50 & 1.45 \\
\midrule
\multicolumn{6}{l}{\bf OWL-vit model fine-tuned on SODA} \\
\midrule
OWL-vit & 0.21 & 0.35 & 0.50 & 0.44 & 1.44 \\
Ours (Task \textbf{3}) & 0.23 & 0.37 & 0.51  & 0.47 & 1.58\\
\midrule
\multicolumn{6}{l}{\bf OWL-vit model fine-tuned on SODA + LVIS} \\
\midrule
OWL-vit & 0.21 & 0.35 & 0.52 & 0.51 & 1.59 \\
Ours (Task \textbf{3}) & 0.23 & 0.39 & \bf0.53  & \bf0.53 & \bf1.67\\
\bottomrule
\end{tabular}}
\caption{\small Detection results on the CODA test set. AP-Com (AP-Common): average precision (AP) on common classes, AP-A (AP-agnostic): AP in the class agnostic setting. AR-A and AR-Cor: average recall for class agnostic and corner case detection respectively. Task 3: Open vocabulary ability (mixture of seen and unseen classes) }
 \label{tab:det:coda}
\end{table}

\noindent\textbf{LVIS experiments.} We provide average precision results for all 4 Tasks on the LVIS mini-validation set~\cite{kamath2021mdetr} in Table~\ref{tab:det:lvis}. For context, we report methods competitive with OWL-vit, but highlight the key point of comparison is with respect to the base model that exhibits pre-defined class names. We also provide comparison to prompt learning, using four learnable context words initialised as `a photo of a', and identical training parameters to ours. 

When learning names for all classes, we achieve consistent performance gains for all class groups in both zero-shot (\mbox{Task \textbf{1}}: adapting to a new dataset, +1.3 AP) and fine-tuned settings (Task \textbf{2}: post training class adjustment, +5.7 AP). Larger gains are observed for the fine-tuned model, which can be attributed to the improved bounding box regression component, facilitating the classification task. Gains are more substantial for rare classes (+2.2 AP (zero-shot), +10.8 AP (fine-tuned)), which can be attributed to the removal of name ambiguity and the fact that these names are less common in pre-training datasets, leading to potentially poorer mapping with image content (\eg `arctic' shoe class). In contrast, frequent classes remain mostly stable (+0 AP (zero-shot), +1.2 (fine-tuned)), highlighting the quality of image to name mapping for this group. We provide more evidence of this in Section~\ref{sec:interp}.

Open vocabulary experiments (Task \textbf{3}), where class names are only learned on base classes, show modest gains in the zero-shot setting (+ 0.5 AP), but large improvements on a fine-tuned model (+ 4.5 AP), notably highlighting the importance of the localisation task. Due to the multi-label nature of LVIS, base and rare classes are not competing, facilitating the open vocabulary task. Learning base class names additionally reduces ambiguity with rare class names, improving rare class performance. 

Sequential learning experiments (Task \textbf{4}; training on base then rare classes) achieve comparable performance to directly training on all classes (35.0 AP (seq.) \vs 35.6 AP (all classes)), further highlighting how well suited our approach is for sequential adaptation. LVIS is a highly imbalanced dataset. While our 10\% subsampling strategy reduces class imbalance, it is not entirely eliminated. 
This is evidenced by the improved rare class performance in the sequential setting (+3.3 (zero-shot), +2.1 (fine-tuned)). Fine-tuning the entire OWL-vit model on all classes ($100\%$ of the data) further evidences the impact of this imbalance and our model's potential to address long tail issues. By reducing data requirements (only $10\%$ achieves equivalent overall performance; 34.5 AP), we are able to train on a balanced data subset. In both fine-tuned and zero-shot settings, learning class names on $10\%$ of the data achieves better or equivalent rare class performance (AP/r) than the fine-tuned model (-0.1 to +15.4 AP/r gains).

Finally, prompt tuning methods struggle to achieve good Task \textbf{1} performance in the zero-shot setting (-0.9 AP), but obtain better gains after fine-tuning (+2.5 AP). We note that prompt tuning performance is overall poorer than our strategy (-2.2 zero shot, -3.2 fine-tuned \vs ours all), which we hypothesise is partially linked to OWL-vit pre-training using fixed templates, and the diversity of object detection contexts. 

\noindent\textbf{CODA experiments, Task 3.} Experimental results on the CODA 2.0 dataset are reported in Table~\ref{tab:det:coda}. Following the setup proposed in~\cite{li2022coda}, we report average precision (AP) on common classes (AP-Com), class agnostic AP on all classes (AP-A), class agnostic average recall on all classes (AR-A) and on corner classes (AR-Cor). We compare to state of the art self-driving models, trained in a semi-supervised setting on the SODA dataset~\cite{li2022coda}. The zero-shot OWL-vit model is not trained on any self-driving data, therefore achieves an overall poorer common class performance (0.10 AP-Com). However, the open-vocabulary setting allows this model to achieve very strong corner classes detection (0.45 AR-Cor). Our method largely improves performance for all metrics, with an overall performance (1.45 Sum) exceeding that of self-driving models (1.41 Sum), including our OWL-vit model fine-tuned on SODA (1.44 Sum). Fine-tuning OWL-vit on SODA, and a mixture of SODA and LVIS training data further improves results (1.44 and 1.59 Sum, respectively), and our strategy consistently increases all metrics (1.58 and 1.67 Sum, respectively). Common class performance remains below the reference models (-0.07 AP-Com), which we conjecture can be attributed to our relatively small self driving training data (only $10k$ images).

\begin{table}[t]
\centering
\resizebox{0.9\linewidth}{!}{
\begin{tabular}
{K{\dimexpr0.18\linewidth-2\tabcolsep-\arrayrulewidth\relax}|
K{\dimexpr0.35\linewidth-2\tabcolsep-\arrayrulewidth\relax}|
K{\dimexpr0.5\linewidth-2\tabcolsep-\arrayrulewidth\relax}|
K{\dimexpr0.3\linewidth-2\tabcolsep-\arrayrulewidth\relax}
}
\toprule
Dataset & Original name & Closest to new name (embedding similarity) & Example image \\
\midrule
\multicolumn{3}{l}{Apperance based Correction}    \\
\midrule
LVIS & arctic & boot (0.46), ski boot (0.30) & 
\raisebox{-0.8\totalheight}{\includegraphics[height=15mm]{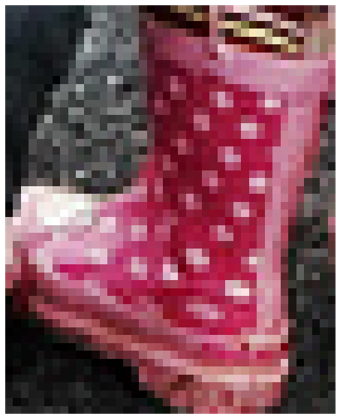}} \\
CODA & tricycle & cart (0.62), rickshaw (0.67) &  \raisebox{-0.8\totalheight}{\includegraphics[height=15mm]{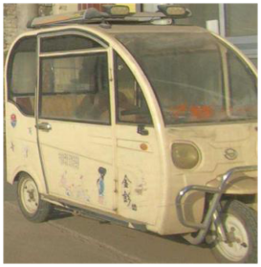}} \\
CODA & misc & waterwheel (0.47), wheel (0.45) & \raisebox{-0.8\totalheight}{\includegraphics[width=\linewidth]{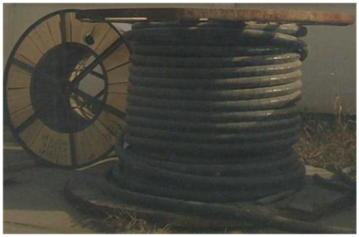}} \\
\midrule
\multicolumn{3}{l}{British to American English Correction}    \\
\midrule
\multirow{4}{*}{LVIS}  & clothes hamper      & \multicolumn{2}{l}{laundry basket (0.52), shopping basket (0.39)}      \\
&wall socket         & \multicolumn{2}{l}{outlet (0.62), power outlet (0.60)}\\
&garbage             & \multicolumn{2}{l}{trash can (0.47), trashcan (0.45)}   \\
&postbox             & \multicolumn{2}{l}{mailbox (0.47), telephone booth (0.46)} \\
&trousers            & \multicolumn{2}{l}{clothes (0.42), pants (0.41)}\\
\midrule
\multicolumn{2}{l}{Object type first}   \\     
\midrule
\multirow{4}{*}{LVIS}  & bell pepper         & \multicolumn{2}{l}{vegetable (0.55), pepper (0.51), bell pepper (0.51)} \\
& bridal gown         & \multicolumn{2}{l}{gown (0.57), dress (0.57), bridal gown (0.55)}       \\
& horse buggy         & \multicolumn{2}{l}{horse-drawn vehicle (0.66), horse buggy (0.64) }     \\
& horse carriage      & \multicolumn{2}{l}{horse-drawn vehicle (0.68), horse carriage (0.65) }  \\
& celery              & \multicolumn{2}{l}{green vegetables (0.50), celery (0.47) }             \\
& cymbal              & \multicolumn{2}{l}{musical instrument (0.55), cymbal (0.46)}        \\
\bottomrule
\end{tabular}%
}
\caption{\small Interpretability results: changes to original class names.} 
\label{tab:interp}
\end{table}

\begin{figure}[t]
\centering
\includegraphics[width=0.55\linewidth]{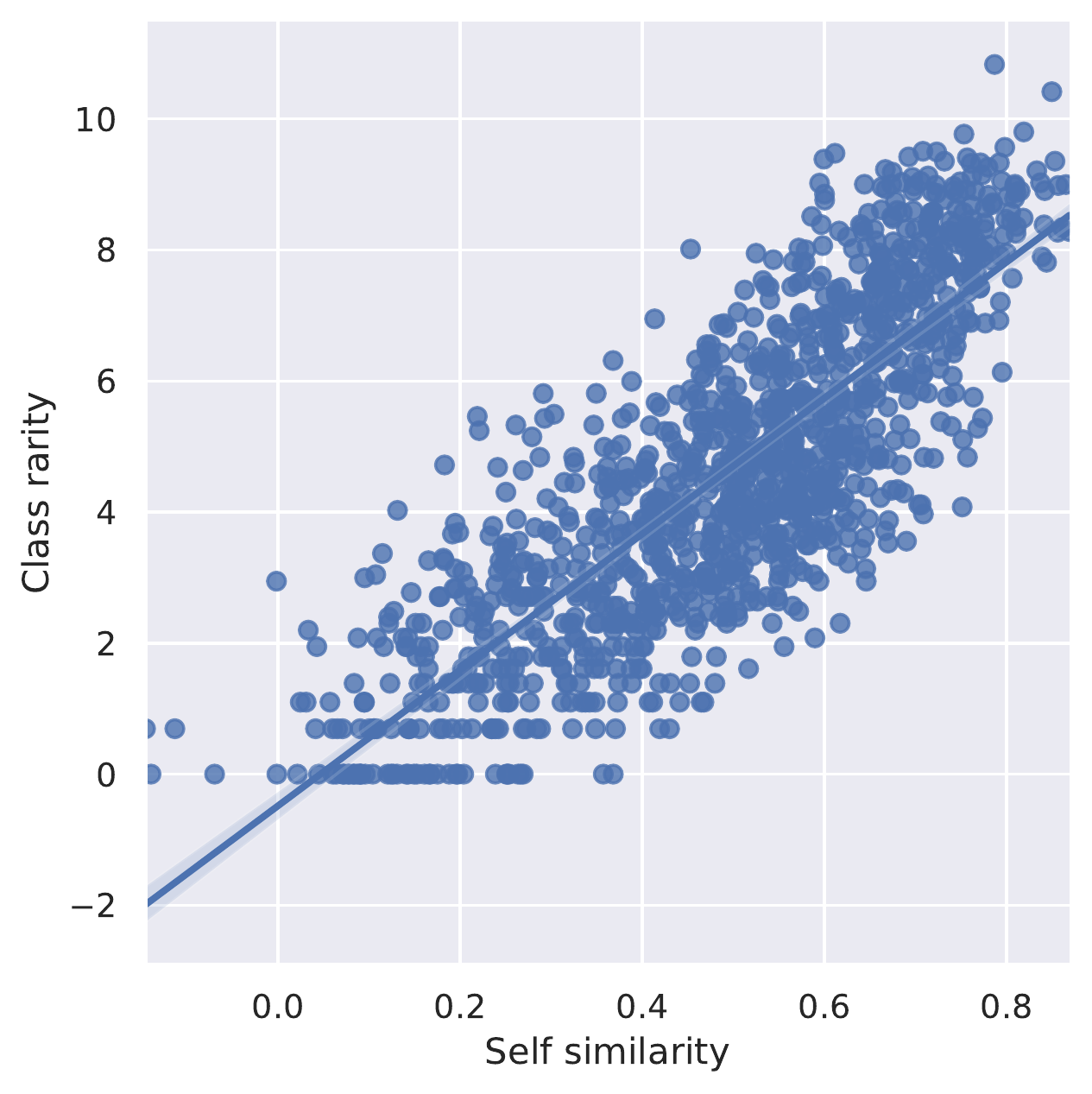}
\captionof{figure}{\small Correlation between class rarity (log scale) and embedding similarity between learned and pre-defined class names. LVIS dataset. 
}
\label{fig:corr}
\end{figure}

\subsection{Interpretability}
\label{sec:interp}

One key advantage of the proposed approach is a beneficial effect on model interpretability; the method allows evaluation of the manner in which class names are modified, based on image content. 
Here, we aim to evaluate how object classes were renamed in our object detection experiments. This experiment can usefully inform us about model biases, sources of classification errors, potential mislabelling errors and the suitability of handcrafted names. 
For this purpose, we build a reference vocabulary of $3031$ object names, using the following dataset class names (filtered for duplicates): LVIS ($1203$ classes), CODA ($43$ classes), Objects$365$~\cite{objects365} ($365$ classes), ImageNet ($1000$ classes), and the THINGS~\cite{things} database ($1854$ concepts).  

Individual word embeddings may not have very valuable meanings in terms of interpretability, especially considering that certain class names comprise multiple tokens. As a result, we compare learned representations to our reference vocabulary at the output of the text transformer encoder, using the shared prompt template context: `a photo of a'. More formally, for each word $\textbf{e}^l$, in our list of learned embeddings, we compute the sentence representation $f(t(\textbf{e}^l))$, where $t(\textbf{e}^l)$ is the sentence constructed using our prompt prefix and word embedding $\textbf{e}^l$. We repeat the procedure for each instance in our vocabulary, using the model's pre-trained word embeddings $\textbf{e}^v$, and obtain $f(t(\textbf{e}^v))$. We then simply compute the cosine distance between all pairs of learned and reference word representations.

We carry out this process for models trained on the LVIS dataset (base model fine-tuned on base classes), and on the CODA 2.0 dataset (zero-shot model). In Table~\ref{tab:interp}, we show examples of classes that were renamed towards a semantic meaning closer to image content, and highlight some labelling errors. ``Arctic'' and ``misc'' are evidence of non representative class names. The ``tricycle'' class appears to have been mislabelled, with labelled vehicles looking closer to ``rickshaws'' and ``carts''. We additionally report two interesting trends: 1) the model tends to adjust word embeddings from British English representations to American English, where the latter is expected to be the dominant language, on which large scale CLIP models were trained; 2) fine-grained category representations tend to be closer to supercategory word types (\eg~vegetable, musical instrument), suggesting that the model also leverages cross-class similarities towards the learning of representations for rare classes. The first observation notably highlights our method's potential for cross-lingual model adaptation: given a dataset in one language, and a model trained in another language, we can learn optimal class names adapted to the pre-trained language model without any additional language translation efforts.

In Figure~\ref{fig:corr} we highlight the relationship connecting class rarity and embedding similarity between learned and original class names, on the LVIS dataset. As previously conjectured in Section~\ref{sec:exp:det}, there is a clear correlation between these variables, highlighting how multi-modal models also suffer from common long tail related issues (\ie poorer semantic mapping between rare words and their corresponding visual appearance). 
Additional results are available in our supplementary materials.

\section{Conclusion}
\label{sec:conclusion}

\looseness=-1
This work focuses on the concept of class name and resulting impact on performance of vision-language recognition models. Complementary to prompt tuning, which seeks to optimise the context of text queries, we propose to automatically learn optimal class names from visual content. 
Our solution can easily be integrated in classification and object detection models, achieving significant gains, with respect to the base model, in all settings.  We additionally demonstrate promising interpretability properties by showing how class semantic meanings were adjusted, providing data and model analysis properties.

\looseness=-1
Learning class names can be viewed as an alternative to linear probing. The former approach, in contrast to probing, maintains key properties inherent to vision-language models (\eg open-vocabulary). One observed limitation of our approach, in common with linear probing, is relatively weaker performance, in comparison with prompt tuning and image queries, under very low data regimes (\ie 1-2 shots). Exploring a combination of learned text and image queries makes for a promising future avenue. While learning class names reduces dependency on prompt context, we still observe improvements using engineering prompts, in particular in open-vocabulary settings. As such, our method has the potential to further benefit from combination with prompt tuning techniques~\cite{zhou2022learning,zhou2022conditional}, where prompts can provide key domain adaptation properties. Last but not least, learning class names has the potential to benefit class agnostic detection, via clustering and learning cluster semantics. 

{\small
\bibliographystyle{ieee_fullname}
\bibliography{egbib}
}

\clearpage
\appendix 

\section{Additional LVIS experiments}

\begin{figure}[t]
\centering
\includegraphics[width=\linewidth]{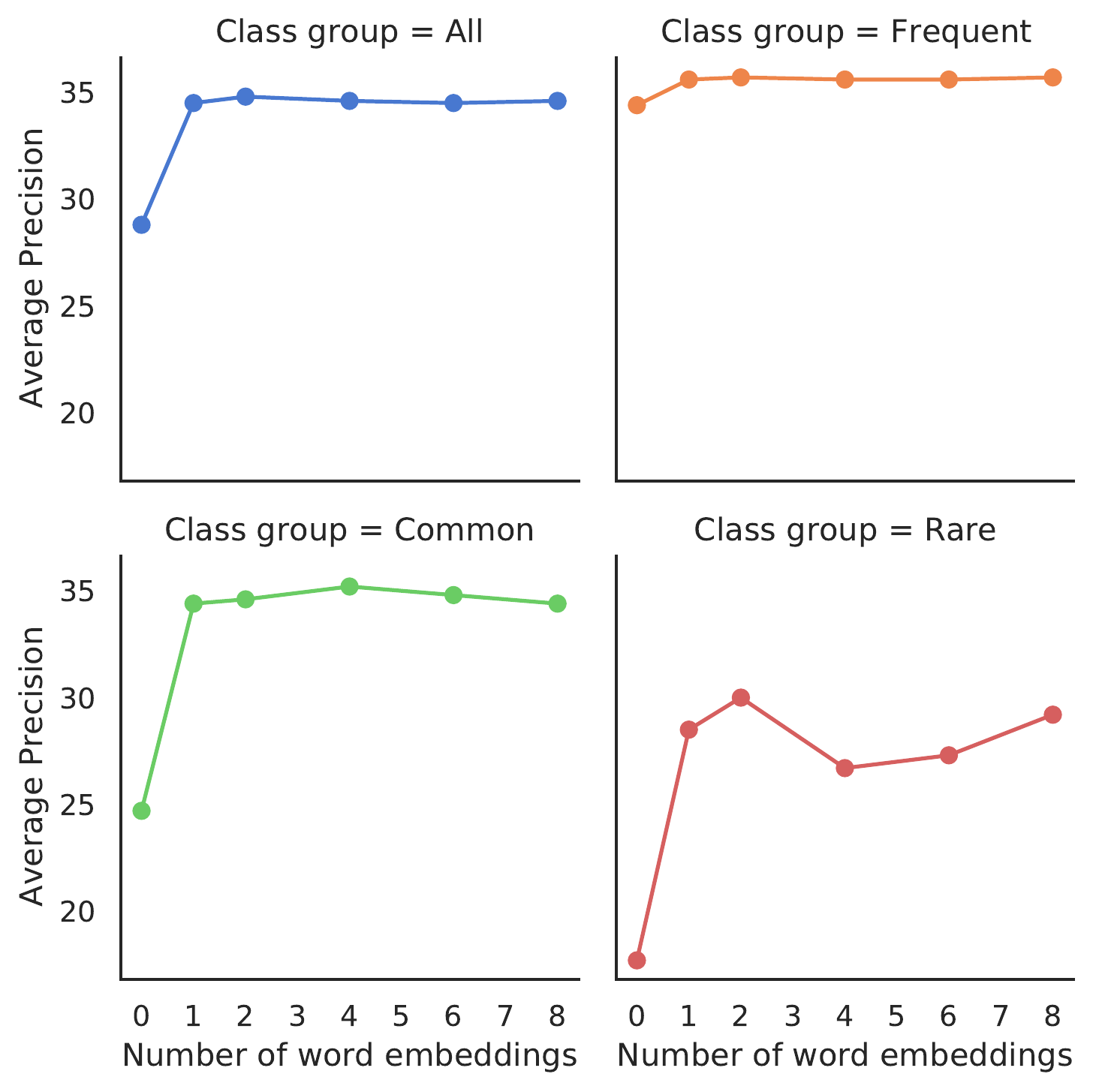}
\caption{Influence of the number of word embeddings learned, per class, on the average precision for class groups: all classes (top-left); frequent classes ($x{>}100$, top-right); common classes ($100{\ge}x{>}10$, bottom-left); and rare classes ($10{\ge}x{\ge}1$, bottom-right), where $x$ pertains to available class samples in each case.}
\label{fig:nemb}
\end{figure}

We provide additional experiments on the LVIS dataset~\cite{lvis} to further analyse the behaviour of our method. Firstly, we evaluate the impact of learning more than one word embedding per class; \ie~rather than using the query \mbox{$t_i= [\text{a photo of a }] + [\text{pl}^i_1] + [\text{.}]$}, we consider multiple placeholders, namely: 
\mbox{$t_i= [\text{a photo of a }] + [\text{pl}^i_1] + \cdots + [\text{pl}^i_m] + [\text{.}]$}, where $m$ is the number of word embeddings to learn. We run experiments for $m=[0,1,2,4,6,8]$, with $m=0$ corresponding to the base model and report results in Figure~\ref{fig:nemb}. It can be observed that a large gain in performance is obtained when replacing original words ($m=0$) with learnable ones ($m=1$), with overall average precision remaining stable as $m$ increases. Rare classes obtain the largest performance gains, while frequent classes have the smallest increase, as discussed in the main manuscript Sec.~4.2. We observe that rare class performance is more unstable as the number of embeddings increases, which can be attributed to the limited available training data, with increased parameter count increasing related overfitting risks. Based on these observed results, we can see that a single word embedding per class is sufficient to achieve good performance, and additional parameters yield no to very limited improvements.

Secondly, we provide results on the full LVIS validation set, alongside results from the detection specific prompt learning technique DETPRO~\cite{du2022learning}, in Table~\ref{tab:lvis_fullval}. We highlight that DETPRO uses VILD~\cite{vild} as a base model (pre-trained on base classes only), and is therefore not directly comparable with our results. We observe similar trends as seen on the mini-validation set, with our strategy achieving large performance gains compared to the original model (+2.9 AP ours-base, +4.9 AP Ours-all). We also observe largest gains for common and rare classes (+7.2 AP/c, AP/r, Ours-all). We note that, compared to the model trained on the full dataset, performance gains are more limited than those observed on the mini-validation set. We conjecture that this could be attributed to the larger number of rare classes, with very few training samples available, (the mini-validation set is not comprised of all classes). DETPRO gains on the rare classes additionally highlight the potential for prompt learning to work in conjunction with our approach, by improving performance in the open-vocabulary setting.

Finally, one additional advantage of our method is that we can achieve strong performance using only $10\%$ of the LVIS training data, when learning class names. This notably allows us to boost performance on rare classes, which are typically penalised by the long tail distribution of the training dataset. As further analysis we provide, in Table~\ref{tab:lvis_10vs100}, results using our balanced subset of $10\%$ of the training data, and the entire, imbalanced, training dataset. We can see that overall performance is largely increased using $100\%$ of the training data, and notably better than a model fine-tuned on the whole dataset in all categories (+1.3 AP). However, we observe that our model trained with a balanced subset achieves stronger performance on rare classes (+3.1 AP/r), confirming the supposed advantage of using a balanced dataset. Disentangling whether performance gains, on both frequent and common classes, are predominantly due to the additional data or the positive bias favouring these class groups is a promising avenue for further investigation.  

\begin{table}[t]
\centering
\begin{tabular}{l|c|c|c|c}
\toprule
Method         & AP   & AP/f & AP/c & AP/r \\
\midrule
\rowcolor{sgmgray} VILD-base      & 27.5 & 31.9 & 27.5 & 17.4   \\
\rowcolor{sgmgray} DETPRO-base    & 28.4 & 32.4 & 27.8 & 20.8 \\
\midrule
OWL-vit-base   &  24.4    &  30.6   &   21.4   &  18.0 \\
Ours-base      &   27.5   &   31.6   &   27.6   &  18.0    \\
Prompt-all     &    26.1  &   31.2   &   24.5   & 18.7 \\
Ours-all       &   29.3   &   32.0   &  \bf28.6    &  \bf25.2   \\
OWL-vit-all     &  \bf30.5    &  \bf35.0    &  \bf28.7    &  24.8   \\
\bottomrule
\end{tabular}
\caption{Average precision detection results on the LVIS full validation set. `-Base' and `-all' indicates the model was trained on base and all classes respectively. `Ours' and `prompt' were trained on the OWL-vit-base model and $10\%$ of the training data, OWL-vit-all is trained with 100\%.}
\label{tab:lvis_fullval}
\end{table}

\begin{table}[t]
\centering
\begin{tabular}{l|c|c|c|c}
\toprule
Method         & AP   & AP/f & AP/c & AP/r \\
\midrule
OWL-vit-base   &  28.8    &  34.4   &   24.7   &  17.7 \\
Ours-all 10\% &  34.5   &   35.6   &   34.4   &   \bf28.5    \\
Ours-all 100\% & \bf35.8    &   \bf38.5   &   \bf34.8   &   25.4  \\
OWL-vit-all  & 34.5  &   \bf38.5   &   33.2   &   19.1   \\
\bottomrule
\end{tabular}
\caption{Average precision detection results on the LVIS mini-validation set: training class names with 10\% of the data \vs 100\%. }
\label{tab:lvis_10vs100}
\end{table}

\section{Detailed classification results}

In this section, we provide detailed, per dataset, results for our classification experiments. Results for model adaptation and open vocabulary experiments are reported in Table \ref{tab:supp:classif:base2new}. In addition to baselines discussed in the main manuscript, we provide, for completeness, results using the CoCoOp method~\cite{zhou2022conditional}. As CoCoOp is significantly more computationally expensive, we only provide results reported in~\cite{zhou2022conditional} that match our experimental setting (in contrast, all CoOp experiments were reproduced locally using the official code and parameters). CoCoOp is an extension of the CoOp method that introduces an image feedback loop using a learnable so-called meta-network to condition prompt learning. We can see that CoCoOp struggles more at Task 1 (adapting to new datasets) that both CoOp and Ours (-2.1\% average accuracy on base classes), achieves stronger open-vocabulary generalisation than CoOp (+7.6\% average accuracy, new classes) but worse than ours and CLIP (-2.51\%). Interestingly, reported average results on all classes outperform our method (+0.49\% average accuracy on all classes), we note the small margin by which our method is outperformed, despite the fact that CoCoOP uses a much more complex and expensive image feedback mechanism. Incorporating a similar feedback loop has the potential to increase the performance of our approach as well.

Detailed results for sequential adaptation are found in Table~\ref{tab:supp:classif:seq}. One notable result from sequential adaptation is that we achieve the most significant gains for datasets with technical class names such as Stanford cars, which can be attributed to the fact that class names are (all) learned independently (\ie there is no mixture of learned and handcrafted technical names).

\begin{table*}[t]
\centering
\resizebox{0.9\linewidth}{!}{\begin{tabular}{l|c|c|c|c|c||c|c|c|c|c||c|c|c|c|c}
\toprule
Dataset & CLIP* & CoOp & CoCoOp\dag & Ours & Ours* & CLIP* & CoOp & CoCoOp\dag & Ours & Ours* & CLIP* & CoOp & CoCoOp\dag & Ours & Ours* \\
\midrule
 & \multicolumn{5}{c||}{\bf Base classes} & \multicolumn{5}{c||}{\bf New classes} & \multicolumn{5}{c}{\bf All classes} \\
 \midrule
Eurosat & 56.4 & \bf92.7 & 87.9 & 92.1 & 91.3 & \bf63.9  & 52.7 &  60.04 & 59.8 & \bf63.9 & 47.7  & 55.2 & - & 51.7 & \bf59.5 \\
Stanford Cars & 63.3 & 77.3 & 70.49 & \bf80.6 & \bf80.6 & \bf75.0 & 61.9 & 73.49 &\bf75.0 & \bf75.0 & 65.3 &  \bf65.7 & - & 63.1 & 63.1 \\
Flowers 102 & 72.2 & 97.5 & 94.87 & 98.0 & \bf98.5 & \bf77.9 & 63.4 & 71.75 & 77.1 & \bf77.9 & 71.4 & 71.5 & - &81.5 & \bf82.6  \\
Oxford Pets & 91.3 & 93.7 & \bf 95.20  & 93.4 & 93.7 & 96.9 & 93.3 & \bf97.69 & 96.9 & 97 &  \bf89.1 & 88.2 & - & 84.5 & 88.0 \\
UCF 101 & 70.6 & 84.2 & 82.33 & 84.1 & 84.6 & \bf77.4 & 57.3 & 73.45 &73.9 & \bf77.4 & 66.7 & 64.6  & - & 70.4 & \bf71.4 \\
Aircraft & 27.6 & 40.5 & 33.41 & \bf44.0 & 43.7 & \bf36.1 & 23.5 & 23.71 & 33.3 & \bf36.2 & 24.6 & 26.3 & - & 19.4 & \bf27.7 \\
DTD & 53.4 & 80.0 & 77.01 & 80.3 & 80.1 & \bf60.3 & 41.5 & 56.00& \bf60.0 & \bf60.0 & 44.5 & 49.5 & - &54.8& \bf57.6 \\
ImageNet & 72.4 & \bf76.4 & 75.98 & 74.7 & 74.7 & 68.1 & 68.2 & \bf70.43 & 68.1 & 68.1 & 66.7 & \bf68.9 & - & 67.7 & 67.7 \\
Caltech 101 & 97.0  & 98.1 & 97.96 & 97.9 & 97.9 & \bf94.0 & 88.8 & 93.81 &\bf94.0 & \bf94.0 & 93.0 & 91.3 & - &\bf93.3  & \bf93.3 \\
Food 101 & 90.1 & 88.0 & \bf90.70 & 86.5 & 86.7 & \bf91.3 & 83.9 & \bf 91.29 & 91.0 & \bf91.3 & \bf86.1 & 79.9 & - & 81.2 & 81.5 \\
Sun 397 & 69.4 & \bf80.6 & 79.74 & 79.1 & 79.1 & 75.5 & 63.2 & \bf76.86 & 75.5 & 75.5 & 62.6 & 62.5 & - &\bf64.3 &\bf64.3  \\
\midrule
Average & 69.4 & 82.6 & 80.47 & 82.7 & 82.8 & \bf74.2 & 64.1 & 71.69 & 73.1 & \bf74.2 & 65.2 & 65.7 & \bf69.19 &66.5 & 68.7 \\
\bottomrule
\end{tabular}}
\caption{\small Detailed results for base to new classification accuracy. * Manually engineered prompt templates. \dag results copied from~\cite{zhou2022conditional}.}
\label{tab:supp:classif:base2new}
\end{table*}

\begin{table}[t]
\resizebox{\linewidth}{!}{\begin{tabular}{l||l|l|l|l||l|l|l|l}
\toprule
Dataset & CLIP* & CoOp & Ours & Ours* & CLIP* & CoOp & Ours & Ours* \\
\midrule
 &  \multicolumn{4}{c||}{\bf New classes} & \multicolumn{4}{c}{\bf All classes} \\
 \midrule
Eurosat & 63.9  & 92.3 &  92.9 & \bf93.4 & 47.7  & 65.3 & 75.1 & \bf76.5 \\
Stanford Cars & 75.0 & 82.7 & \bf90.8 & \bf90.8 & 65.3 & 71.8 & \bf80.9 & \bf80.9 \\
Flowers 102  & 77.9 & 97.0 & \bf98.5 & 98.4 & 71.4 & 81.3 & \bf96.3 & \bf96.3 \\
Oxford Pets & 96.9 & \bf97.6 & 97.4 & 97.3 &  \bf89.1 & 88.5 & 86.2 & 87.5 \\
UCF 101  & 77.4 & 87.2 & \bf87.7 & 87.5 & 66.7 & 74.6 & \bf80.3 & 80.0 \\
Aircraft  & 36.1 & 53.6 & \bf62.4 & \bf62.4 & 24.6 & 32.1 & 30.8 & \bf35.6 \\
DTD  & 60.3 & 75.6 & \bf76.0 & 75.9 & 44.5 & 54.9 & \bf67.1 & 66.8 \\
ImageNet  & 68.1 & \bf72.7 & 71.7 & 71.7 & 66.7 & \bf70.8 & 69.3 & 69.3 \\
Caltech 101 & 94.0 & \bf95.9 & 95.6 & 95.6 & 93.0 & \bf94.2 & 93.9 & 93.9 \\
Food 101 & 91.3 & \bf91.4 & 89.6 & 89.7 & \bf86.1 & 84.9 & 82.1 & 81.6 \\
Sun 397 &  75.5 & 82.1 & \bf82.4 & \bf82.4 & 62.6 & 71.0 & \bf71.6 & \bf71.6 \\
\midrule
Average &  74.0 & 84.4 & \bf85.9 & \bf85.9 & 65.2 & 71.7 & 75.7 & \bf76.3\\
\bottomrule
\end{tabular}}
\caption{\small Detailed classification accuracy for sequential training. * manually engineered prompt templates.}
\label{tab:supp:classif:seq}
\end{table}

\section{CLIP engineered templates}

We provide the list of manually engineered templates for CLIP zero-shot classification, (mentioned in our main paper, Sec.~4.1), 
in Table~\ref{tab:templates}. We highlight that four datasets use our standard template (``a photo of a'') and three datasets exhibit distinctly disparate sentence templates (namely Eurosat, DTD and UCF 101).

\begin{table}[t]
\begin{tabular}{l|l}
\toprule
Dataset         & Template \\
\midrule
Eurosat & a centered satellite photo of [CLASS]. \\
Stanford Cars & a photo of a [CLASS]. \emph{(default)} \\
Flowers 102 & a photo of a [CLASS], a type of flower. \\
Oxford Pets & a photo of a [CLASS], a type of pet. \\
UCF 101 & a photo of a person doing [CLASS]. \\
Aircraft & a photo of a [CLASS], a type of aircraft. \\
DTD & [CLASS] texture. \\
ImageNet & a photo of a [CLASS]. \emph{(default)} \\
Caltech 101 & a photo of a [CLASS]. \emph{(default)} \\
Food 101 & a photo of [CLASS], a type of food. \\
Sun 397 & a photo of a [CLASS]. \emph{(default)} \\
\bottomrule
\end{tabular}
\caption{List of engineered templates used for CLIP zero-shot classification, for each of the eleven datasets. }
\label{tab:templates}
\end{table}

\section{Interpretability}
\label{sec:supp:interp}

\begin{figure*}[t]
\centering
\includegraphics[width=\linewidth]{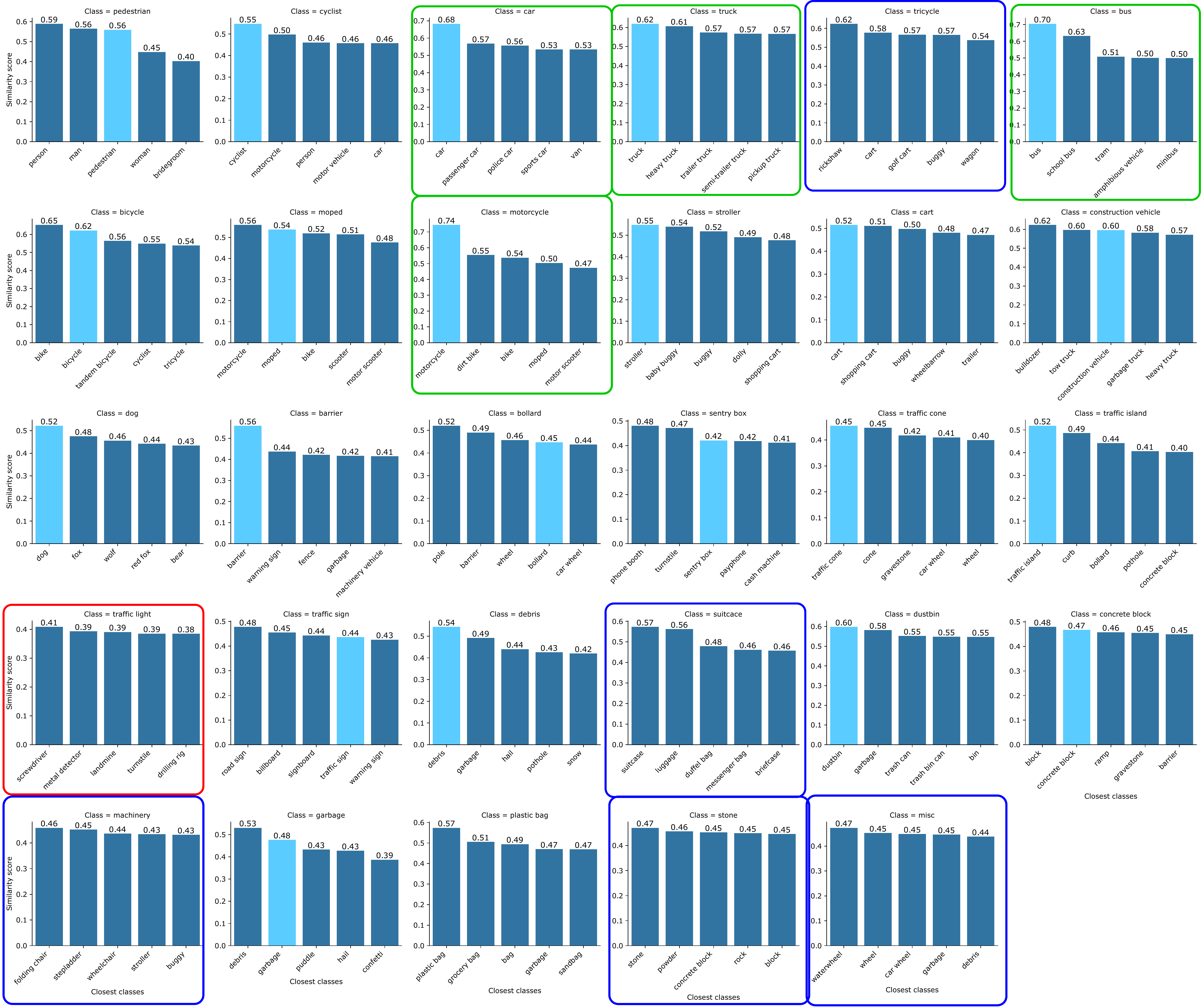}
\caption{Interpretability results on the CODA 2.0 dataset using the \textbf{zero-shot base model} (see main paper, Sec.~4.3). 
Light blue bars: highlight the similarity between new word embeddings and original class name embeddings; highlighted classes (green boxes): strong similarity with original class name (${>}0.6$, closest word), highlighted classes (blue boxes): original name not included within top 5 most similar classes, highlighted classes (red boxes): modified name semantic meaning markedly different from original.}
\label{fig:intCODA}
\end{figure*}

\begin{figure*}[t]
\centering
\includegraphics[width=\linewidth]{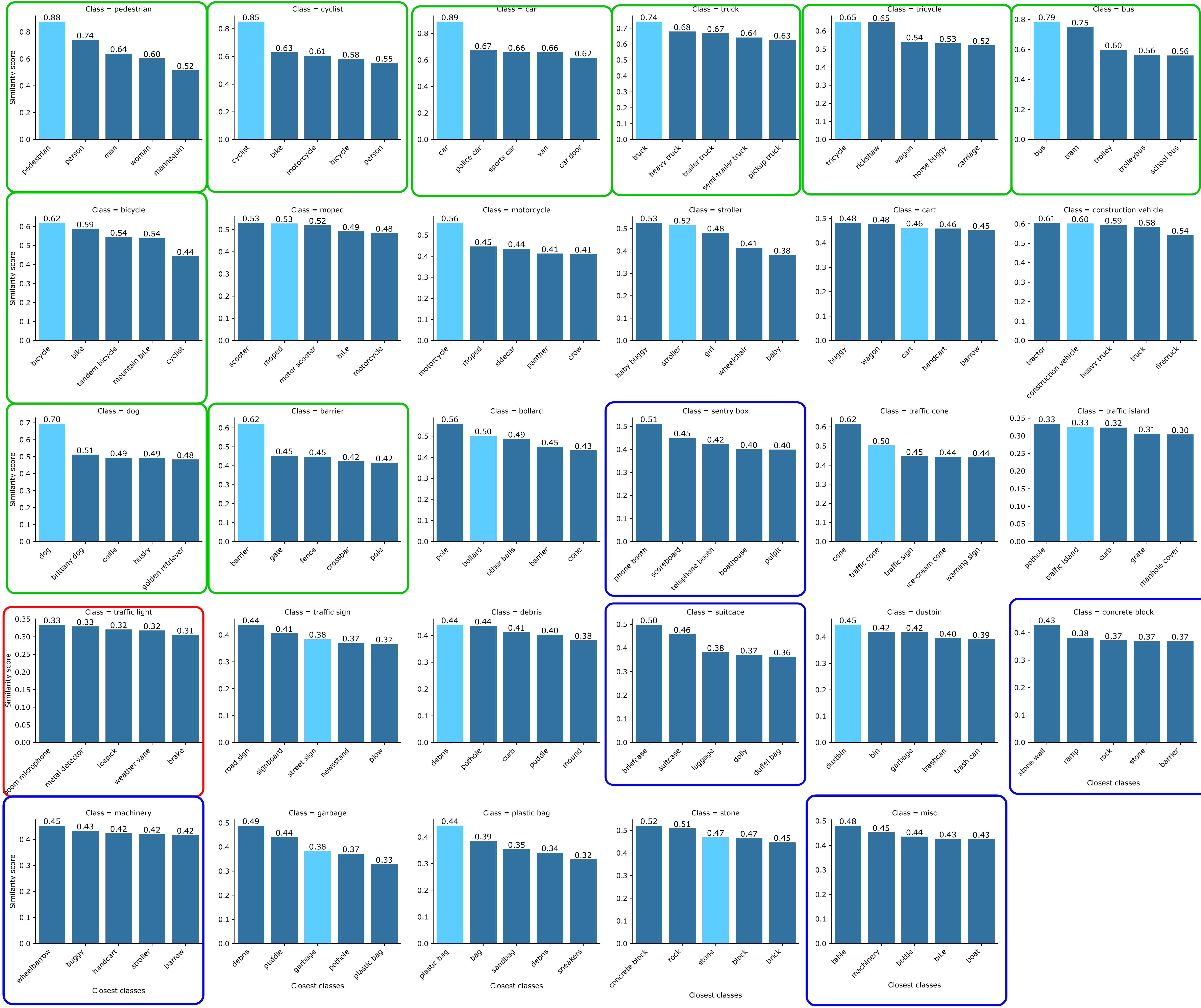}
\caption{Interpretability results on the CODA 2.0 dataset on the \textbf{base model fine-tuned on LVIS and SODA} (see main paper, Sec.~4.3). 
Light blue bars: highlight the similarity between new word embeddings and original class name embeddings; highlighted classes (green boxes): strong similarity with original class name ($>0.6$, closest word), highlighted classes (blue boxes): original name not included within top 5 most similar classes, (red boxes): modified name semantic meaning markedly different from original.}
\label{fig:intCODAFT}
\end{figure*}

We further provide additional results with regards to our interpretability experiments (see main paper Sec.~4.3), for the CODA 2.0 dataset~\cite{li2022coda}. In Figures~\ref{fig:intCODA} and~\ref{fig:intCODAFT}, we illustrate how class names were modified for all $29$ classes in the validation set, for the zero-shot and fine-tuned model (LVIS + SODA~\cite{han2021soda10m}), respectively. On the zero-shot models, we can see that self-driving specific terms (\eg~pedestrian), map to more common terms with similar meaning (\eg~person). We highlight in particular the tricycle class, which is mapped to `rickshaw', a more appropriate term for the visual content available in the dataset. We additionally note that class names were more substantially modified, as the highest observed self-similarity is $0.74$ (motorcycle). In comparison, the fine-tuned model shows much higher class name stability, especially with regards to common classes (first row). This provides insight into what was learned during the fine-tuning process, and the similarity between class names of SODA and CODA 2.0 datasets (\eg~tricycle class).

\begin{figure}[t]
    \centering
    \begin{subfigure}[t]{0.1\textwidth}
        \centering
        \includegraphics[width=\textwidth]{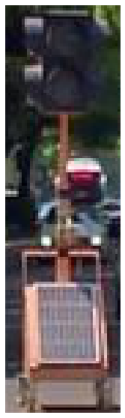}
    \end{subfigure} 
    \begin{subfigure}[t]{0.1\textwidth}
        \centering
        \includegraphics[width=\textwidth]{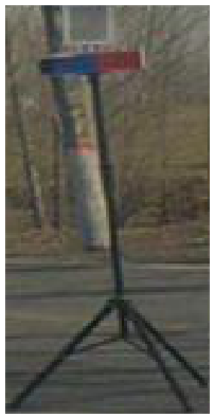}
    \end{subfigure} 
    ~ 
    \begin{subfigure}[t]{0.1\textwidth}
        \centering
        \includegraphics[width=\textwidth]{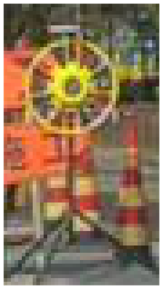}
    \end{subfigure}
    ~ 
    \begin{subfigure}[t]{0.1\textwidth}
        \centering
        \includegraphics[width=\textwidth]{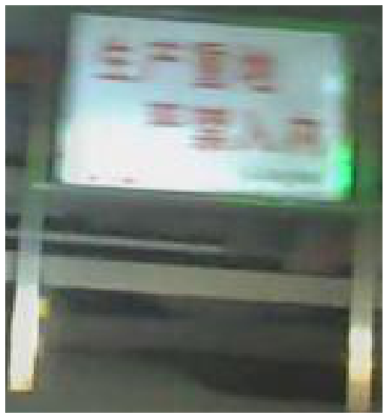}
    \end{subfigure} \\
    \caption{Examples of training samples in the traffic light category.}
    \label{fig:TL}
\end{figure}

We further note large differences in semantic meaning between the learned representations of the `misc.'~and `machinery' classes, when comparing both models. This suggests that these classes do not comprise representative components and should be further separated, potentially via a clustering strategy. Finally, we point out how the new embedding for the traffic light class is poorly adapted to the class' original meaning. This can be attributed to the fact that the data available for this class only comprises mobile traffic lights, in addition to mislabelled samples (see Figure~\ref{fig:TL} for examples of training samples). As such, the visual content is highly different from learned mappings between the traffic light term and standard image content. This highlights that we can easily identify failure modes, potentially allowing, in cases of overfitting or poor training examples, to correct new words of poor quality (\eg use the original word embeddings). 

\section{Visual examples}

\begin{figure*}[t!]
    \centering
    \begin{subfigure}[t]{0.3\textwidth}
        \centering
        \includegraphics[width=\textwidth]{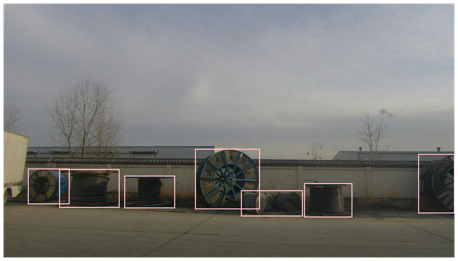}
    \end{subfigure}
    ~ 
    \begin{subfigure}[t]{0.3\textwidth}
        \centering
        \includegraphics[width=\textwidth]{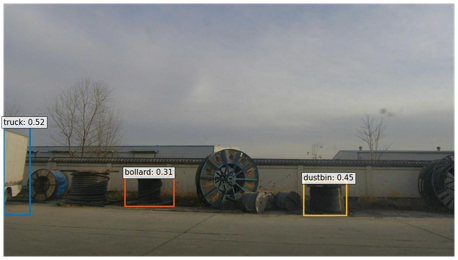}
    \end{subfigure}
    ~ 
    \begin{subfigure}[t]{0.3\textwidth}
        \centering
        \includegraphics[width=\textwidth]{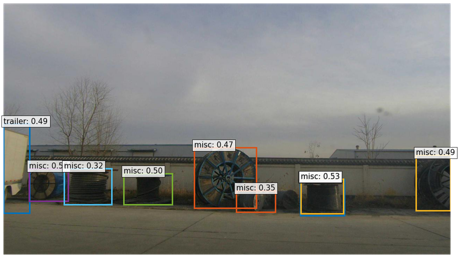}
    \end{subfigure} \\
    \begin{subfigure}[t]{0.3\textwidth}
        \centering
        \includegraphics[width=\textwidth]{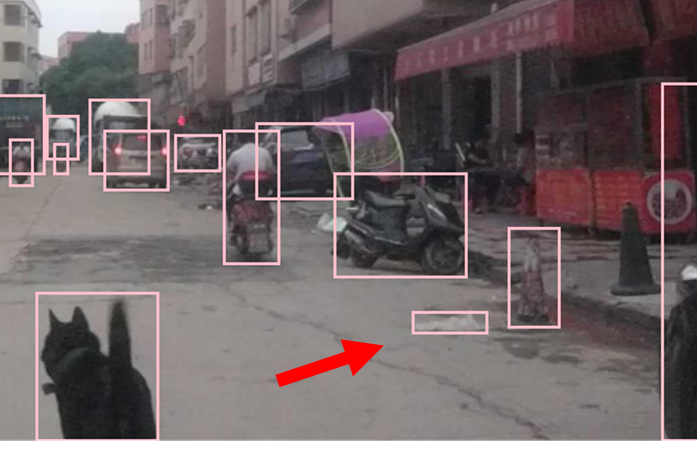}
    \end{subfigure}
    ~ 
    \begin{subfigure}[t]{0.3\textwidth}
        \centering
        \includegraphics[width=\textwidth]{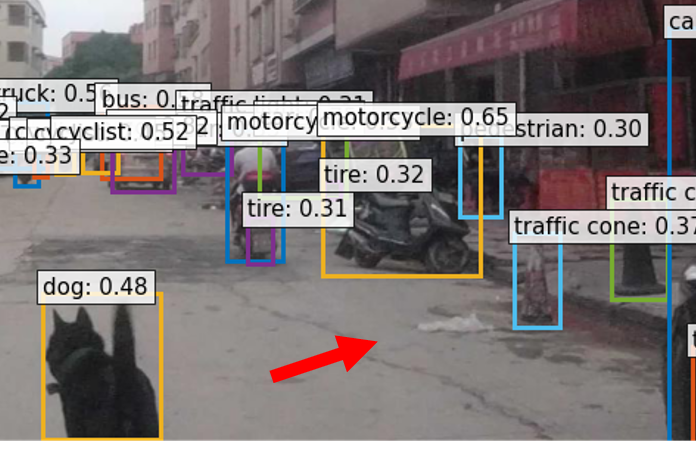}
    \end{subfigure}
    ~ 
    \begin{subfigure}[t]{0.3\textwidth}
        \centering
        \includegraphics[width=\textwidth]{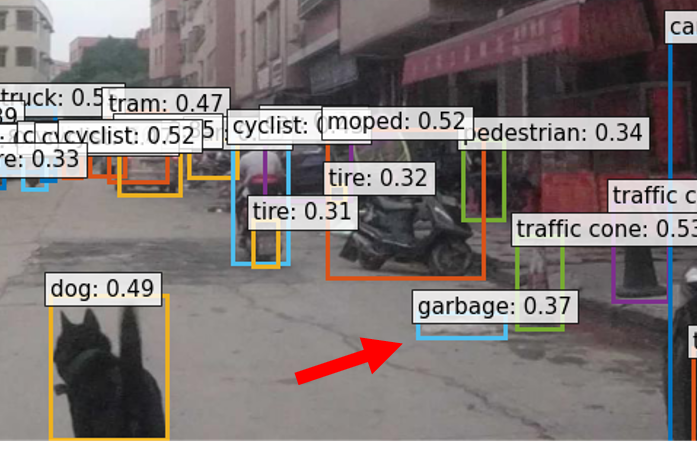}
    \end{subfigure} \\
    \begin{subfigure}[t]{0.3\textwidth}
        \centering
        \includegraphics[width=\textwidth]{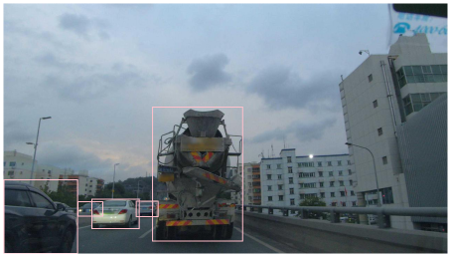}
    \end{subfigure}
    ~ 
    \begin{subfigure}[t]{0.3\textwidth}
        \centering
        \includegraphics[width=\textwidth]{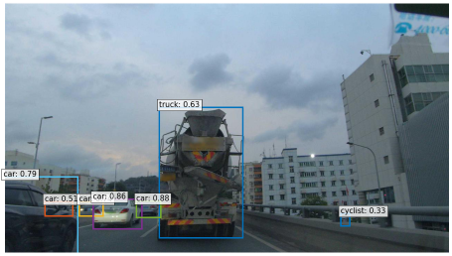}
    \end{subfigure}
    ~ 
    \begin{subfigure}[t]{0.3\textwidth}
        \centering
        \includegraphics[width=\textwidth]{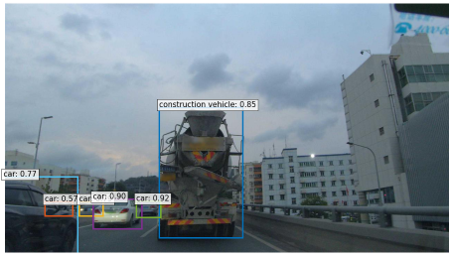}
    \end{subfigure} \\
    \begin{subfigure}[t]{0.3\textwidth}
        \centering
        \includegraphics[width=\textwidth]{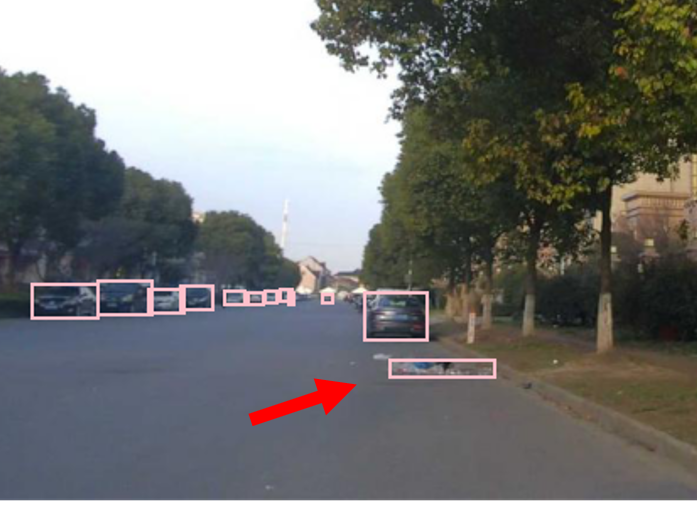}
    \end{subfigure}
    ~ 
    \begin{subfigure}[t]{0.3\textwidth}
        \centering
        \includegraphics[width=\textwidth]{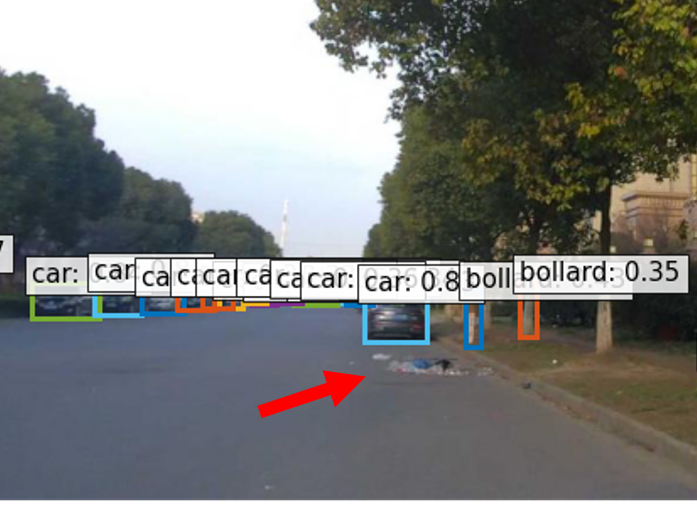}
    \end{subfigure}
    ~ 
    \begin{subfigure}[t]{0.3\textwidth}
        \centering
        \includegraphics[width=\textwidth]{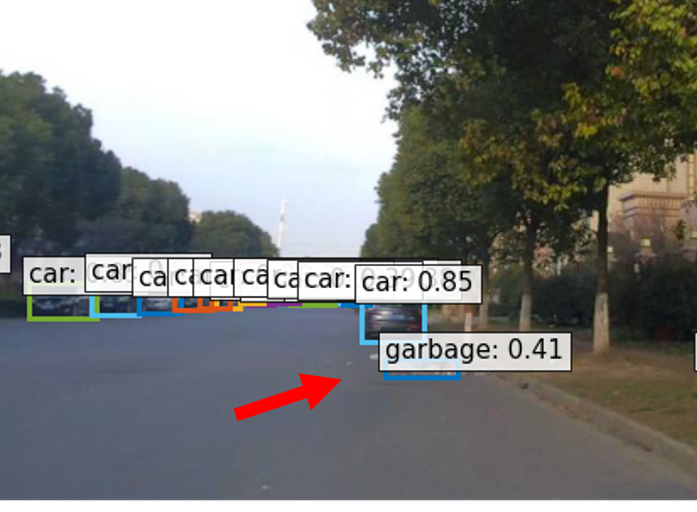}
    \end{subfigure} \\
    \begin{subfigure}[t]{0.3\textwidth}
        \centering
        \includegraphics[width=\textwidth]{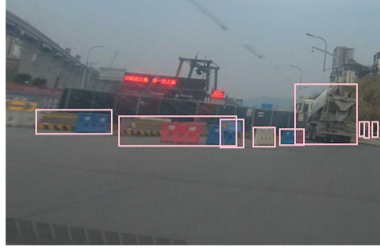}
    \end{subfigure}
    ~ 
    \begin{subfigure}[t]{0.3\textwidth}
        \centering
        \includegraphics[width=\textwidth]{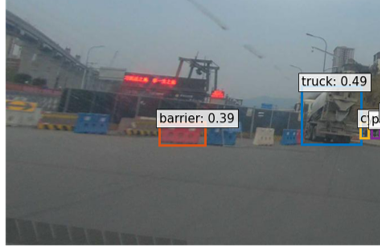}
    \end{subfigure}
    ~ 
    \begin{subfigure}[t]{0.3\textwidth}
        \centering
        \includegraphics[width=\textwidth]{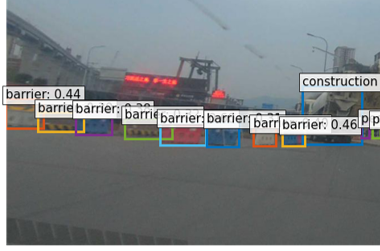}
    \end{subfigure} \\
    \begin{subfigure}[t]{0.3\textwidth}
        \centering
        \includegraphics[width=\textwidth]{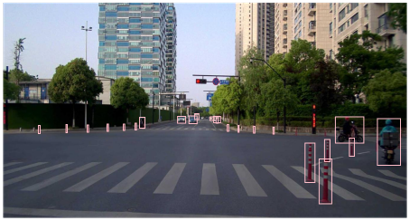}
        \caption{Ground truth bounding boxes}
    \end{subfigure}
    ~ 
    \begin{subfigure}[t]{0.3\textwidth}
        \centering
        \includegraphics[width=\textwidth]{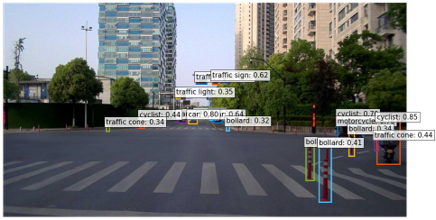}
        \caption{OWL-vit model}
    \end{subfigure}
    ~ 
    \begin{subfigure}[t]{0.3\textwidth}
        \centering
        \includegraphics[width=\textwidth]{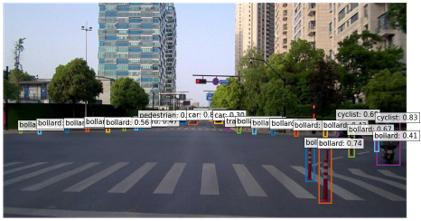}
        \caption{Ours}
    \end{subfigure} \\
    \caption{Visual results from the CODA 2.0 dataset, comparing our approach to the base model, fine-tuned on LVIS + SODA. Our approach exhibits improved performance on corner-case classes.}
    \label{fig:vis}
\end{figure*}

In Figure~\ref{fig:vis}, we provide visual examples of improvements to the object detection task for the CODA 2.0 dataset. We compare our fine-tuned model (LVIS + SODA) to performance after learning class names, and provide the class agnostic ground truth as reference. We highlight how our model is able to recognise instances of the misc.~class, identifying construction vehicles (\vs truck category, improving common class performance), and overall displays a stronger ability to identify corner cases. 

\section{Implementation}

Implementation in MindSpore will be made available at  \url{https://gitee.com/mindspore/models/tree/master/research/cv/}.


\end{document}